# The use of entropy to measure structural diversity

L. Masisi[1], V. Nelwamondo[2] and T. Marwala[1]
[1]School of Electrical & Information Engineering Witwatersrand University, P/Bag 3, Wits, 2050, South Africa
[2]Graduate School of Arts and Sciences, Harvard University, GSAS Mail Center Child 412, 26 Everett Street Cambridge, Massachusetts, 02138 USA

*Abstract—* **In this paper entropy based methods are compared and used to measure structural diversity of an ensemble of 21 classifiers. This measure is mostly applied in ecology, whereby species counts are used as a measure of diversity. The measures used were Shannon entropy, Simpsons and the Berger Parker diversity indexes. As the diversity indexes increased so did the accuracy of the ensemble. An ensemble dominated by classifiers with the same structure produced poor accuracy. Uncertainty rule from information theory was also used to further define diversity. Genetic algorithms were used to find the optimal ensemble by using the diversity indices as the cost function. The method of voting was used to aggregate the decisions.**

## I. Introduction

There is still an immense need to develop robust and reliable classification of data. It has become apparent that as opposed to using one classifier an ensemble of classifiers performs better [1],[2],[3]. This is because a committee of classifiers is better than one classifier. However one of the question that arises is that, how do we measure the integrity of these committee in generalizing. The popular method that is used do gain confidence from the generalization ability of an ensemble is of inducing diversity within the ensemble. This therefore calls for a form of a method of measuring the diversity of the ensemble. Methods have been implemented to relate ensemble diversity with ensemble accuracy[4],[5],[6]. These methods use the outcomes of the individual classifiers of the ensemble to measure diversity [7],[8]. This means that diversity is induced by different training methods, popular ones being, boosting and bagging.

This paper deals with the measure of structural diversity of an ensemble by using entropy measures. Diversity is induced by varying the structural parameters of the classifiers [9]. The parameters of interest include the activation function, number of hidden nodes and the learning rate. This study therefore does not take into consideration the outcome of the individual classifiers to measure diversity but the individual structure of the classifiers of the ensemble. One of the statistical measures of variance such as the Khohavi variance method has already been used to measure structural diversity of an ensemble [9]. This study aims to find a suitable measure of structural diversity by using methods adopted in ecology and also use the concept of uncertainty adopted in information theory to better understand the ensemble diversity. The entropy measures are therefore aimed at bringing more knowledge to how diversity of an ensemble relates with the ensemble accuracy. However this study will only focus on three measures of diversity, Shannon, Simpson and Berger Parker to quantify structural diversity of the classifiers.

Shannon entropy has found its fame in information theory as it is used to measure the uncertainty of states[10]. In ecology Shannon is used to measure the diversity indices of species, however in this study instead of the biological species the individual classifiers are treated as species [11]. For example, if there are three species of different kind, two of the same kind and one of another kind, then that would replicate three MLP's of different structural parameters.

The relationship between the classification accuracy and the entropy measures is attained by the use of genetic algorithms by using accuracy as the cost function [9]. There are a number of aggregation schemes such as minimum, maximum, product, average, simple majority, weighted majority, Naïve Bayes and decision templates to name a few [12], [13]. However for this study the majority vote scheme was used to aggregate the individual classifiers for a final solution. This paper includes a section on the background, Species and the Identity Structure (IDS), Renyi Entropy, Shannon entropy measure, Simpson Diversity Index, Berger Parker Index, The neural network parameters, Genetic Algorithms (GA), The model, The data used, Results and discussion and then lastly the conclusion.

## II. Background

Shannon entropy has been used in information theory to quantify uncertainty [10]. The meaning or implication of information is not dealt with in this paper. However this paper aims to use similar concepts. The more information one has the more certain one becomes [10], likewise we can postulate that the more diverse something is the more uncertain we become in knowing its decision or outcome. This can be accredited to the use of Shannon entropy to quantify uncertainty. Entropy measures have been used to compute species population diversity [14], however in this paper a committee of classifiers with different parameters is considered as a committee of species.

## III. Species and the Identity Structure

The ensemble of the classifiers was then treated as species when viewed in the perspective of ecology or as population in statistics. However before the ecological methods can be applied in giving an indication of structural diversity, it was important that the classifiers have a unique identity. This was due to the fact that the ensemble was composed of classifiers with different machine parameters such as the hidden nodes, learning rates and the type of the activation function used. The Identity Structure (IDS) was converted to a binary string so as to mimic a gene type unique for the classifiers.

$$IDS = \begin{bmatrix} Activation\ function \\ Number\ of\ hidden\ nodes \\ Learning\ rate \end{bmatrix}$$

Five learning rates were considered and three activation functions just as in[9]. The number of hidden nodes was between 7 and 21. They were made larger than the attributes so as to have classifiers that could generalize well and then less than 21 so as to reduce the computational costs. The learning rates considered were: 0.01, 0.02, 0.03, 0.04, 0.05 and the activation functions were: The sigmoid, linear, and the logistic. This paper is a continuation of [9].

In this paper there was no need to convert the identity into a binary string since the entropy measure only looks at the machines which are different. The individual classifiers forming the ensemble were given different numbers as according to their identity. Defining an identity for each machine is necessary so as to have a unique identity of the classifiers within the ensemble. This will intern enable the use of the uncertainty measure on the ensemble.

### IV. RENYI ENTROPY

The Renyi entrony equation can be decomposed into Shanon, Berger parker and Simpson's entropies. This choice of the entropy measure is determined by the value of the alpha($\propto$) variable, see equation (1)[15].

$$H_\propto = \frac{ln(\sum P_i^\propto)}{1-\propto} \quad (1)$$

Where: $P$ is the proportion of an item i.

The diversity measures can be found by, Shannon ($\propto \to 1$), Simpson's ($\propto \to 2$) and the Berger Parker ($\propto \to \infty$).

### V. SHANNON ENTROPY

Shannon Entropy in information theory is perceived as the measure of uncertainty. If the states of the process cause the process after 10 iterations to give a series of ones, then one would be certain of the next preceding information. However if the states are diverse then we become uncertain of the outcome. Having an ensemble of classifiers which are all the same, would imply that if one of them were to classify an object. Then with high probability all of them would classify the same object alike. However the more diverse the ensemble become the more uncertain one is of the overall decision of the ensemble. This analogy was used to relate diversity and uncertainty in this paper. In information theory the uncertainty is seen as bits per symbol[10]. The uncertainty can be partially explained from the following equation, by using logs.

$$u_i = -\log(p_i) \quad (2)$$

Where, $p = 1/M$ is the probability that any symbol appears, which means in this case $p$ is the probability of choosing any classifier within the ensemble and $u$ is the uncertainty. Equation (2) tends to infinity likewise if $p$ tends to 1. Shannon's general formula for uncertainty, see Equation (3) when $\propto$ tends to 1 from equation (1).

$$H_1 = -\sum_{i=1}^{M} P_i log(P_i) \quad (3)$$

Where, M is the total number of the classifiers.

The maximum of Equation (3) occurs when the structural diversities of the classifiers are equally likely. This means when $P_i$ = 1/M for all classifiers, substituting this into Equation (3) will result in, *log (M),* this is normally perceived as species richness in ecology[11].

For this study the Shannon entropy was normalized between 0 and 1 by dividing Equation (3) by log(M) the maximum possible diversity index. That means a 1 will mean the largest uncertainty (high diversity index) of the system and then 0 would mean no uncertainty.

### VI. SIMPSON'S DIVERSITY INDEX

When taking $\alpha$ to 2, the Renyi entropy approximates to, see equation (4):

$$H_2 = -log(\sum_{i=1}^{n} P_i^2) \quad (4)$$

It is based on the idea that that the probability of any two individuals drawn at random from a large ecosystem belonging to different species is given by $\sum p_i^2$ [16]. The inverse of this expression is taken as the biodiversity index, that means $H_2$ increases with the evenness of the distribution which is the diversity index in this case. A 1 will represent more diversity and zero no diversity. The normalization was done removing the *log* and then by using, 1 - $H_2$ so that as the evenness increases so does the diversity index.

### VII. BERGER PARKER INDEX

The Renyi entropy approximates to, see Equation (5) [16], when taking $\alpha$ to infinity:

$$H_\infty = \frac{1}{p_i} \quad (5)$$

$H_\infty$ gives the equivalent number of equally abundant species with the same relative abundance as the most abundant species in the system [16]. The Berger Parker index only considers the relative dominance of the most popular species, ignoring all the other species. The Berger Parker index was normalized between 0 and 1 by dividing $H_\infty$ by 21 the total number of the classifiers within the ensemble.

## VIII. THE NEURAL NETWORK (NN) PARAMETERS

The entropy measures were based on the structural parameters of the neural network. See Figure 1 of the Multi Layered Perceptron (MLP):

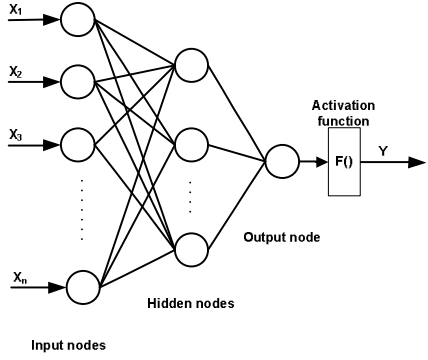

Figure 1: The MLP structure showing the inputs, the layers and the activation function

An MLP is mainly used to map the inputs to the outputs. The activation functions used in this paper include the linear, sigmoid and the Softmax. They can be found at the outer layer of the MLP and hence are called the activation functions. The MLP is also well known for the hidden nodes, biases and weights. In this study the MLP is used as a classifier, an ensemble of classifiers is made however with different structural parameters. This means the classifiers would not all classify the same.

That means that if we had the same classifiers with the same structural parameters then we would be certain of the classification of the rest if we had information about any one of the classification of one classifier. That means the degree of uncertainty increased as the classifiers were more different. Hence it should be apparent that the uncertainty is not induced by different training schemes. The output of the NN is perceived as the probability, see Equation (6) [17], which describes the output of the neural network. This means in implementing the method of voting we would be dealing with the probability of each classifier in either agreeing or rejecting the decision taken by other classifiers.

$$y_k = f_{outer}\left(\sum_{j=1}^{M} w_{kj}^{(2)} f_{inner}\left(\sum_{i=1}^{d} w_{ji}^{(1)} x_i + w_{jo}^{(1)}\right) + w_{ko}^{(2)}\right) \quad (6)$$

Where, $f_{outer}$ and $f_{inner}$ are the activation functions at the output layer and at the hidden layer respectively, M is the number of the hidden units, d is the number of input units, $w_{ji}^{(1)}$ and $w_{kj}^{(2)}$ are the weights in the first and second layer respectively when moving from input i to hidden unit j, and $w_{jo}^{(1)}$ is the biases of the unit j.

Classifiers of different structural parameters were created. This was done so as to induce structural diversity on the ensemble.

## IX. GENETIC ALGORITHMS (GA)

GA are evolutionary algorithms that aim to find a global solution to a certain problem domain. The GA makes use of the principles of evolutionary biology, such as mutation, crossover, reproduction and natural selection [18]. Hence the GA has high capabilities to search large spaces for an optimal solution. The search process of the GA includes:

1. Generation of a population of offspring, mostly interpreted as chromosomes

2. A cost or evaluation function, that is used to control the whole process of selection and rejection of chromosomes via a process of mutation a and crossover functions.

3. This process will continue until the fittest chromosome is attained or the termination of the process can be defined other than the one mentioned

In this study the evaluation function is the ensemble accuracy, the GA searches for a group of 21 classifiers that would minimize the cost function. The number of classifiers within the ensemble was abstractly chosen. That means an ensemble that will produce the targeted accuracy. The GA searches through already trained 120 classifiers. In essence the GA evolves the artificial machines (classifiers) to attain this goal.

## X. THE MODEL

The model describes the use of GA in selecting the 21 classifiers so as to provide knowledge of how the accuracy of the ensemble relates with the uncertainty of the ensemble. Figure (2) that illustrates the use of 120 classifiers in attaining an optimal ensemble for classification. A method of voting is used to aggregate the individual decisions of the classifiers within the ensemble.

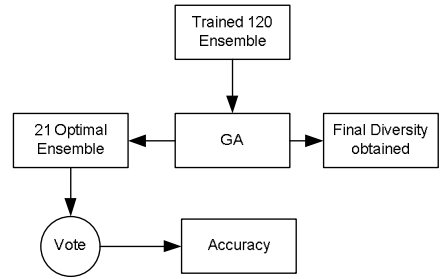

Figure 2: The method used to optimize the 21 classifiers of the 120 classifiers

The evaluation function is composed of two variables, the ensemble accuracy and the targeted accuracy $T_{acc}$. Equation (7) is used as the evaluation function. The ensemble was chosen to have 21 classifiers, the number was made odd so that there would not be a tie during voting, and chosen abstractly.

$$f_{GA} = -(Acc - T_{acc})^2 \qquad (7)$$

Where: $f_{GA}$ is the evaluation function, $Acc$ is the accuracy of the 21 classifiers and $T_{acc}$ is the targeted accuracy.

The GA tries to optimize the accuracy of the ensemble evaluation function by finding its maximum. Equation (7) is a parabolic function that has an optimal point at zero. This in term would mean that the desired accuracy would be reached. GA was first optimized by first searching the target values which the it could attain. These were then the targeted accuracy values for the cost function. This was done so as to reduce the computational cost since the search space will be minimized.

### A. Ensemble Generalization

The classification accuracy of the ensemble was attained by using a method of voting to aggregate the individual decision of the classifiers. For every classification done on the data sample, the number of correct classification was counted. See Equation (8) for calculating the classification accuracy of the ensemble.

$$Acc = \frac{n}{N} \qquad (8)$$

Where: $n$ and $N$ is the number of the correct classification and the total number of the sample data to be classified, respectively.

The NN was taken as a probability measure output values equal to or larger than 0.5 were taken as a 1 and output values less than 0.5 were considered as zero.

## XI. THE DATA

Interstate conflict data obtained was used for this study. There are 7 attributes and one output, see table 1 for the data input features. The output is binary, a 0 for no conflict and a 1 for conflict. There are a total of 27,737 cases in the cold war population. The 26,846 are the peaceful dyads year and 875 conflict dyads year [19]. This shows clearly that the data are complicated for training a neural network. The data were then doubled as according to the training, validation and testing data sets. For this study the significance of the data was not considered. This data was used to demonstrate the system behavior in regards to the entropy measures.

Table 1: The interstate conflict data

| Inputs | Values |
|---|---|
| Allies | 0-1 |
| Contingency | 0-1 |
| Distance | Log10(Km) |
| Major Power | 1-0 |
| Capability | Log10 |
| Democracy | -10-10 |
| Dependency | continuous |

The data was normalized between 0 and 1, to have equal weight of all the features by using Equation (9) so that all features were equally weighted for training.

$$X_{norm} = \frac{x_i - x_{min}}{x_{max} - x_{min}} \qquad (9)$$

Where $x_{min}$ and $x_{max}$ are the minimum and maximum values of the features of the data samples observed, respectively.

## XII. RESULTS AND DISCUSSION

The entropy measures were done on the ensemble of 21 classifiers. These measures are quantified as the diversity indices of the ensembles. See figure 3, 4 and 5 for the diversity indexes and an indication of the structural diversity of the ensembles. These are the results of 11 ensembles as were selected by the GA.

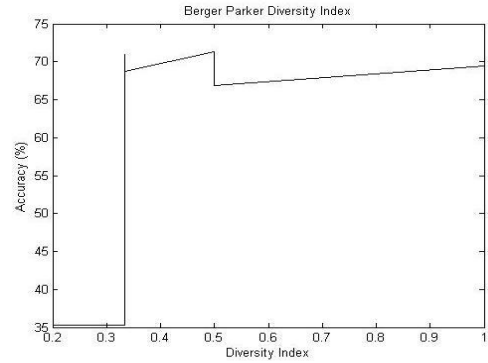
Figure 3: The Berger Parker index of diversity and accuracy

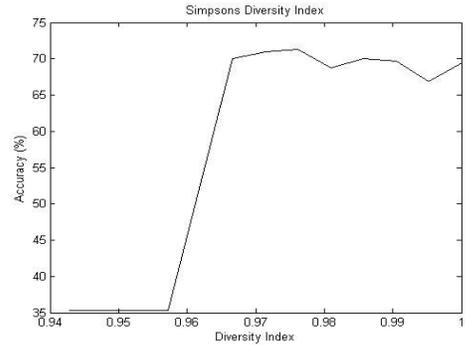
Figure 4: The Simpson's diversity index and accuracy

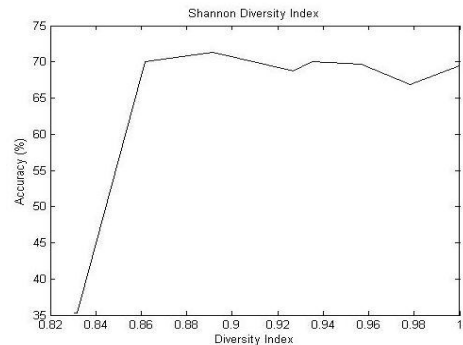
Figure 5: The Shannon diversity index and accuracy

The Shannon diversity index indicate that at very low diversity index the generalization of the ensemble is poor, however as the diversity increases so does the accuracy. There seems to be a high correlation between the Shannon and the Simpson's diversity indices in relation to the classification accuracy, results from the Simpson's measure shows to be more sensitive towards high diversity indices.

The Shannon diversity index and the Simpson's diversity indices have a decreasing accuracy after reaching a peak accuracy level, see figure 4 and 5. This indicates that evenness on the classifiers needs to be limited for good ensembles. Accuracies of up to 71 % where attained. The use of accuracy as a function of Berger Parker diversity measure did not show to be a good function of Berger Parker measure of structural diversity of the ensemble. This can be seen on Figure 3.

## XIII. CONCLUSION

This paper presented the use of entropy based measures to quantify structural diversity. This diversity measures where then compared to the ensemble accuracy. Three measures of diversity indices were compared and it was observed that the ensembles accuracy improved as the structural diversity of the classifiers increased. The other interesting observation was that of the Shannon diversity index when interpreted as the uncertainty measure from the information theory. As the uncertainty of the ensemble increase so did the classification of the ensemble. This implies that having more information of the ensemble might result in poor generalization ability of the ensemble, hypothetically. The method used to compute the results was computationally expensive due to the use of GA. This paper has also showed that Entropy based methods can be used to better understand the ensemble diversity in particular ensemble structural diversity. However the use of measuring structural diversity in building good ensembles of classifiers still remains to be explored.